\documentclass{article}

\usepackage{arxiv}

\usepackage[utf8]{inputenc} % allow utf-8 input
\usepackage[T1]{fontenc}    % use 8-bit T1 fonts
\usepackage{hyperref}       % hyperlinks
\usepackage{url}            % simple URL typesetting
\usepackage{booktabs}       % professional-quality tables
\usepackage{amsfonts}       % blackboard math symbols
\usepackage{nicefrac}       % compact symbols for 1/2, etc.
\usepackage{microtype}      % microtypography
\usepackage{lipsum}		% Can be removed after putting your text content
\usepackage{graphicx}
\usepackage{natbib}
\usepackage{doi}
\DeclareRobustCommand{\bbone}{\text{\usefont{U}{bbold}{m}{n}1}}
\usepackage{amsmath,amssymb}

\title{The NT-Xent loss upper bound}

\author{\hspace{1mm} Wilhelm Ågren \{\texttt{wagren@kth.se}\}\\
    Division of Computational Science and Technology\\
    School of Electrical Engineering and Computer Science\\
    KTH Royal Institute of Technology\\
    SE-100 44 Stockholm, Sweden
}

\date{}

\renewcommand{\headeright}{}
\renewcommand{\undertitle}{An upper bound for average similarity}
\renewcommand{\shorttitle}{}

\hypersetup{
pdftitle={Properties of the NT-Xent loss},
pdfauthor={Wilhelm Ågren},
pdfkeywords={SimCLR, NT-Xent, Self-supervised learning},
}

\begin{document}
\maketitle

\begin{abstract}
    Self-supervised learning is a growing paradigm in deep representation learning, showing great generalization capabilities and competitive performance in low-labeled data regimes. The SimCLR framework proposes the NT-Xent loss for contrastive representation learning. The objective of the loss function is to maximize agreement, similarity, between sampled positive pairs. This short paper derives and proposes an upper bound for the loss and average similarity. An analysis of the implications is however not provided, but we strongly encourage anyone in the field to conduct this.
\end{abstract}


\section{Introduction}
Traditional deep neural networks are becoming larger, deeper, and more complex in order to keep pushing and defining the competitive limits for machine learning. Modern deep learning is mostly based on traditional supervised learning, leveraging massive amounts of labeled data to train models. As such, there is often an abundance of unlabeled data that can not be used in a traditional supervised context. The incentive to research and develop alternative methods has been growing in recent years. Self-supervised learning models and frameworks like CPC\citep{oord2019cpc}, SimCLR \citep{chen2020simclr}, BYOL \citep{grill2020byol}, and SimSIAM \citep{chen2020simsiam} show promising results in low-labeled data regimes, with representations generalizing well in downstream tasks like visual image classification. The mentioned models usually leverage contrastive loss functions to learn co-occuring modalities present in the unlabeled data. These loss functions have already been explored, studied, and evaluated rather extensively, but there does not seem to exist a defined upper bound for the NT-Xent loss used by the SimCLR framework. Therefore, this short papers aims at defining one such boundary.

\section{The SimCLR framework}
\label{sec:SimCLR}
The simple framework for contrastive learning on visual representations (SimCLR) proposed by \citep{chen2020simclr} builds upon properties of noise contrastive estimation (NCE) and is inspired by previous work in the related field. It aims to simplify the process of contrastive representation learning, by not requiring the use of memory banks for sampling nor maximization of mutual information between latent space representations \citep{wu2018memory, oord2019cpc}. Instead, positive pairs are generated by leveraging data augmentations and SimCLR learns representations by maximizing an agreement between the positive pairs. The authors demonstrate that leveraging SimCLR in a low-labeled data regime produces competitive performance and even succeeds in outperforming the AlexNet when fine-tuned on 100$\times$ fewer labels. The framework can be formalized by four main components:
\begin{itemize}
    \item A stochastic data augmentation module $\mathcal{T}$ is used to sample two augmentations $t, t' \in \mathcal{T}$ which generates two augmented views $\tilde{x_i} = t(x), \tilde{x_j} = t'(x)$ of an original datapoint $x$. The generated views are referred to as a positive pair, and are assumed to be correlated since they originate from the same datapoint $x$.
    \item An artificial neural network encoder $f$ which is trained to extract features from the augmented views $\tilde{x_i}$ and $\tilde{x_j}$. Thus, the encoder maps $f(\tilde{x_i}) = h_i$ where $h_i \in {\rm I\!R}^d$ is the feature of interest in the embedding space, which is to be used for the domain-specific downstream tasks.
    \item An artificial neural network projection head $g$ that performs a non-linear mapping of the embedding feature to the latent space, $g(h_i) = z_i \in {\rm I\!R}^m$. The authors suggest to use a simple MLP for projection head, with $ReLU(x) = max(0, x)$ activation functions between layers.
    \item A contrastive loss function $\mathcal{L}$ that is applied on the samples in the latent space. Here, the authors propose the temperature scaled loss NT-Xent which is based on noise contrastive estimation and cross entropy \citep{chen2020simclr}.
\end{itemize}

\section{The NT-Xent loss}
\label{sec:NT-Xent}
The normalized temperature-scaled cross entropy loss (NT-Xent) is designed to maximize an agreement between positive pairs in a mini-batch $\mathcal{MB}$. No explicit negative sampling is performed, instead, $\mathcal{MB}$ is constructed such that $N$ datapoints are sampled and applied to $\mathcal{T}$ to generate positive pairs $\tilde{x_i}$ and $\tilde{x_j}$. This yields $\mathcal{MB} = \{\tilde{x_{1i}}, \tilde{x_{1j}}, ..., \tilde{x_{Ni}}, \tilde{x_{Nj}} \}$, and as such, $|\mathcal{MB}| = 2N$. Thus, given a positive pair, the other $2(N-1)$ datapoints are to be treated as negative samples. The agreement to maximize is the cosine distance of similarity,
\begin{equation}
\label{eq:cosine-similarity}
    sim(z_i, z_j) = \frac{z_i \cdot z_j}{\lVert z_i \rVert \lVert z_j \rVert}
\end{equation}
where $z_i$ and $z_j$ are latent space representations of a sampled positive pair. The total loss for all positive pairs is formalized as
\begin{equation}
\label{eq:NT-Xent}
    \mathcal{L}^{\mathrm{NT-Xent}}= -\frac{1}{N}\sum_{i,j\in \mathcal{MB}}  \mathrm{log}\frac{\mathrm{exp}(sim(z_i, z_j)/\tau)}{\sum_{k=1}^{2N} \bbone_{[k \neq i]} \mathrm{exp}(sim(z_i, z_k)/\tau)}
\end{equation}
where $\tau$ is the temperature scalar, and $\bbone_{[k\neq i]} \in \{0, 1\}$ is an indicator function to indicate that the similarity between the same point, $sim(z_i, z_k)_{[k=i]}$, is not contributing to the normalization. Equation (\ref{eq:NT-Xent}) can be rewritten accordingly by applying the logarithmic rules
\begin{equation}
\begin{aligned}
\label{eq:NT-Xent_alt}
    \mathcal{L}^{\mathrm{NT-Xent}} &= -\frac{1}{N}\sum_{i,j\in \mathcal{MB}} \Big( sim(z_i, z_j)/\tau- \mathrm{log}\sum_{k=1}^{2N}\bbone_{[k\neq i]}\mathrm{exp}(sim(z_i, z_k)/\tau)\Big)\\
    &= \underbrace{-\frac{1}{N}\sum_{i,j\in \mathcal{MB}}sim(z_i, z_j)/\tau}_{\mathcal{L}_{alignment}} + \underbrace{\frac{1}{N}\sum_i^N\mathrm{log}\Big(\sum_{k=1}^{2N}\bbone_{[k \neq i]}\mathrm{exp}(sim(z_i, z_k)/\tau)}_{\mathcal{L}_{distribution}}\Big)
\end{aligned}
\end{equation}
and as such there are two parts of the loss, namely, \textit{alignment} and \textit{distribution}, as proposed by \citep{wang2020loss} and further studied by \citep{chen2020loss}. Furthermore, a similar way to analyze the NT-Xent loss was proposed by Grill et al. in the bootstrap your own latent (BYOL) paper where direct comparisons were made to the proposed SimCLR loss \citep{grill2020byol}. The second term includes summation over the well known \textit{LogSumExponent} (LSE) term which is defined as
\begin{equation}
\label{eq:LSE}
    \mathrm{LSE}(x_1, ..., x_n) = \mathrm{log}\Big(\mathrm{exp}(x_1) + \cdots + \mathrm{exp}(x_n)\Big)
\end{equation}
where $x_{i,k}$ will denote $sim(z_i, z_k)/\tau$, $k\in \{1, ..., 2N\}$, and $n=2N$ to be consistent with the NT-Xent loss notations. It can be proven that LSE is bound accordingly (strictly greater for $n>1$ and strictly lower unless all arguments are equal)
\begin{equation}
\label{eq:LSE_bound}
    max(x_{i,1}, ..., x_{i,n}) \leq \mathrm{LSE}(x_{i,1}, ..., x_{i,n}) \leq max(x_{i,1}, ..., x_{i,n}) + \mathrm{log}(n).
\end{equation}
Thus, using equations (\ref{eq:LSE}) and (\ref{eq:LSE_bound}) one can rewrite equation (\ref{eq:NT-Xent_alt} to produce an upper bound for the average similarity between the positive pairs in $\mathcal{MB}$, accordingly
\begin{equation}
\begin{aligned}
    \mathcal{L}^{\mathrm{NT-Xent}} &= -\frac{1}{N}\sum_{i,j\in\mathcal{MB}}x_{i,j} + \frac{1}{N}\sum_i^N\mathrm{LSE}(x_{i,1}, ..., x_{i,n}) \\
    &\leq -\frac{1}{N}\sum_{i,j\in\mathcal{MB}}x_{i,j} + \frac{1}{N}\sum_i^N\Big( max(x_{i,1}, ..., x_{i,n}) + \mathrm{log}(n)\Big) \\
    &= -\frac{1}{N}\sum_{i,j\in\mathcal{MB}}x_{i,j} + \mathrm{log}(n) + \frac{1}{N}\sum_i^N max(x_{i,1}, ..., x_{i,n})
\end{aligned}
\end{equation}
and by rearranging the expression, multiplying by $\tau$, and returning to the original complete notations, one can get the final expression as
\begin{equation}
\label{eq:avg_sim}
    \frac{1}{N}\sum_{i,j\in\mathcal{MB}}sim(z_i, z_j)\leq \tau\mathrm{log}(2N) - \tau\mathcal{L}^{\mathrm{NT-Xent}} + \frac{\tau}{N}\sum_i^N max(sim(z_i, z_1)/\tau, ..., sim(z_i, z_{2N})/\tau).
\end{equation}

\section{Final remarks}
From equation (\ref{eq:avg_sim}) one can clearly see the three crucial terms that put constraints on and alter how well a SimCLR trained model can learn to position positive pairs closely and aligned in the latent space. As always, the objective is to minimize the loss $\mathcal{L}$ which is done by means of gradient descent. However, since the loss expression includes terms directly related to properties of the learned representation, there is an interplay between updating model parameters $\theta$ and altering where the samples are positioned in the latent space. Hence, analyzing the behavior of the loss by only considering one of the terms is highly theoretical, and might not represent the true behavior. Therefore, this paper only aims at presenting the upper bound for average similarity, but hopefully enabling future work to practically analyze and evaluate what has been proposed.

\section{Acknowledgments}
I would like to present a sincere thanks to Arvind Kumar and Erik Fransén for granting me the opportunity to write a master's thesis on self-supervised learning and magnetoencephalography, which inspired me to write this brief paper. Furthermore, I am very grateful for the supervision that Erik Fransén has provided, which motivated me to keep being creative and passionate about machine learning.

\bibliographystyle{unsrtnat}
\bibliography{references}  %%% Uncomment this line and comment out the ``thebibliography'' section below to use the external .bib file (using bibtex) .

%%% Uncomment this section and comment out the \bibliography{references} line above to use inline references.
% \begin{thebibliography}{1}

% 	\bibitem{kour2014real}
% 	George Kour and Raid Saabne.
% 	\newblock Real-time segmentation of on-line handwritten arabic script.
% 	\newblock In {\em Frontiers in Handwriting Recognition (ICFHR), 2014 14th
% 			International Conference on}, pages 417--422. IEEE, 2014.

% 	\bibitem{kour2014fast}
% 	George Kour and Raid Saabne.
% 	\newblock Fast classification of handwritten on-line arabic characters.
% 	\newblock In {\em Soft Computing and Pattern Recognition (SoCPaR), 2014 6th
% 			International Conference of}, pages 312--318. IEEE, 2014.

% 	\bibitem{hadash2018estimate}
% 	Guy Hadash, Einat Kermany, Boaz Carmeli, Ofer Lavi, George Kour, and Alon
% 	Jacovi.
% 	\newblock Estimate and replace: A novel approach to integrating deep neural
% 	networks with existing applications.
% 	\newblock {\em arXiv preprint arXiv:1804.09028}, 2018.

% \end{thebibliography}

\end{document}